\documentclass[nohyperref]{article}

\usepackage{microtype}
\usepackage{graphicx}
\usepackage{subfigure}
\usepackage{booktabs}
\usepackage{hyperref}
\usepackage[accepted]{icml2022}
\usepackage{times}
\usepackage{latexsym}
\usepackage{multirow}
\usepackage{multicol}
\usepackage{amsmath}
\usepackage{amssymb}
\usepackage{mathtools}
\usepackage{amsthm}
\usepackage[T1]{fontenc}
\usepackage[utf8]{inputenc}

\usepackage{amsmath,amsfonts,bm}

\def\eqref#1{equation~\ref{#1}}

\def\1{\bm{1}}

\DeclareMathAlphabet{\mathsfit}{\encodingdefault}{\sfdefault}{m}{sl}
\SetMathAlphabet{\mathsfit}{bold}{\encodingdefault}{\sfdefault}{bx}{n}

\usepackage{url}
\usepackage{graphicx}
\usepackage[capitalize,noabbrev]{cleveref}
\usepackage{xcolor}
\usepackage{dsfont}
\usepackage[normalem]{ulem}

\begin{document}

\twocolumn[
\icmltitlerunning{Direct Simultaneous Speech-to-Speech Translation with Variational Monotonic Multihead Attention}
\icmltitle{Direct Simultaneous Speech-to-Speech Translation with \\ Variational Monotonic Multihead Attention}


\begin{icmlauthorlist}
\icmlauthor{Xutai Ma}{jhu,meta}
\icmlauthor{Hongyu Gong}{meta}
\icmlauthor{Danni Liu}{maastricht,meta}
\icmlauthor{Ann Lee}{meta}
\icmlauthor{Yun Tang}{meta}
\icmlauthor{Peng-Jen Chen}{meta}
\icmlauthor{Wei-Ning Hsu}{meta}
\icmlauthor{Phillip Koehn}{jhu,meta}
\icmlauthor{Juan Pino}{meta}
\end{icmlauthorlist}

\icmlaffiliation{jhu}{Johns Hopkins University}
\icmlaffiliation{meta}{Meta}
\icmlaffiliation{maastricht}{Maastricht University}

\icmlcorrespondingauthor{Xutai Ma}{xutai\_ma@jhu.edu}


\vskip 0.3in
]
\printAffiliationsAndNotice{}

\begin{abstract}
We present a direct simultaneous speech-to-speech translation (Simul-S2ST) model,
Furthermore, the generation of translation is independent from intermediate text representations.
Our approach leverages recent progress on direct speech-to-speech translation with discrete units,
in which a sequence of discrete representations,
instead of continuous spectrogram features,
learned in an unsupervised manner, are predicted from the model and passed directly to a vocoder for speech synthesis on-the-fly.
We also introduce the variational monotonic multihead attention (V-MMA),
to handle the challenge of inefficient policy learning in speech simultaneous translation.
The simultaneous policy then operates on source speech features and target discrete units.
We carry out empirical studies to compare cascaded and direct approach on the Fisher Spanish-English and MuST-C English-Spanish datasets.
Direct simultaneous model is shown to outperform the cascaded model by achieving a better tradeoff between translation quality and latency. 
\end{abstract}
\section{Introduction}
Simultaneous speech-to-speech (Simul-S2ST) translation is the task of incrementally translating speech in a source language to speech in another language given the partial input only.
It aims at applications where low latency is required,
such as international conferences,
where professional human interpreters usually serve the purpose.
While a fair amount of research has been conducted on simultaneous translation,
including text-to-text~\cite{cho2016can, gu2017learning, ma-etal-2019-stacl, arivazhagan-etal-2019-monotonic, ma2019monotonic},
and speech-to-text~\cite{ren2020simulspeech, ma2020simulmt, ma2021streaming},
only a few works have explored the simultaneous translation in the speech-to-speech setting~\cite{zheng-etal-2020-fluent, sudoh2020simultaneous}
which mostly adopt the cascaded approach.

A cascaded Simul-S2ST system is usually composed of three parts,
a streaming automatic speech recognition (ASR) system which transcribes source speech to source text,
a simultaneous text-to-text system which translates source text to target text in a simultaneous fashion,
and an incremental text-to-speech (TTS) synthesis system to generate target speech in real time.
While cascaded systems can be built on top of the existing simultaneous components, it has several disadvantages.
First, the pipeline of multiple models introduce extra latency.
The latency can either come from the computation time of multiple models,
or the delay caused by their asynchronous processing \cite{zheng-etal-2020-fluent}.
The computation speed asynchrony of different systems can cause undesired latency unless the cooperation between them is enabled with a well designed policy.
Meanwhile, errors can be propagated and accumulated through a pipeline of model components.

Recent efforts on direct S2ST provide a new possibility for Simul-S2ST,
where the intermediate representations, such as source or target texts, are no longer needed.
\citet{jia2019direct} introduced Translatotron model which directly generated speech spectrogram features on the target side.
The gap with a cascaded system is bridged in their followup work Translatotron 2 \citep{jia2021translatotron}.
These approaches show potential when there is more training data. 
However, this approach can be computationally inefficient in real-time,
since generating each frame of speech signal requires a high-dimensional continuous feature vector.
More recently, \citet{lee2021direct} proposed a direct S2ST model,
where a sequence of discrete units instead of spectrogram is directly generated from the translation model.
The discrete units are then passed to a vocoder for target speech synthesis.
The labels of units are self-supervised representations learned from HuBERT~\cite{hsu2021hubert}
The vocoder can be trained separately and run on-the-fly given the discrete units.

Furthermore, we proposed a new simultaneous policy,
variational monotonic multihead attention(V-MMA),
to handle the difficulty of policy learning in speech applications.
While several prior works on monotonic attention based policies~\cite{raffel2017online,arivazhagan-etal-2019-monotonic,ma2019monotonic}
use a closed form estimation on simultaneous alignment during the training time,
we find that such estimation can be biased given long sequences,
which usually happens in speech applications.
Instead, we introduce a variational method to estimate such alignment with less model bias and also less mismatch between training and inference. 

In this work, we propose a direct simultaneous speech-to-speech translation model,
based on recent progress on speech-to-units (S2U) translation~\cite{lee2021direct},
along with a novel simultaneous policy V-MMA.
Our model is featured with two characteristics.
First, the model is simultaneous, and able to generate the target speech based on the partial source speech before obtaining the complete input;
second, the model is independent from intermediate text outputs when generating target speech.
We carry out experiments on the Fisher Spanish-English \cite{post2014fisher} and MuST-C English-Spanish datasets \cite{cattoni2021must} and provide a comprehensive empirical comparison between direct and cascaded models in Simul-S2ST.
\section{Task Formalization}
We first formalize the task of Simul-S2ST,
including the definition of the task and its evaluation.
Let $\mathcal{X}(t)$ and $\mathcal{Y}(t)$ denote the input and output amplitude of the speech signal at a given time $t$.
The durations of input and output speech are $T_\mathcal{X}$ and $T_\mathcal{Y}$.
Assuming the input speech happens at $t=0$ and the system starts generation at $t=T_0$,
a system is defined as simultaneous S2ST system if $T_0 < T_\mathcal{X}$.
In an offline system, $T_0 \ge T_{\mathcal{X}}$. 
Similar to a translation task with text output,
the evaluation for simultaneous S2ST includes two aspects, translation quality and latency.
To measure the translation quality of S2ST, an automatic speech recognition (ASR) system is used to transcribe the generated speech 
and the transcription is then compared with reference translation using metrics such as BLEU.

As for latency of Simul-S2ST systems,  there is a lack of well defined metrics.
One of the most popular evaluation metrics for text-to-text simultaneous translation is average lagging
\cite{ma-etal-2019-stacl}, which was later was extended to speech-to-text translations \cite{ma2020simulmt}.
Given a sequence of text predictions $\mathbf{Y}={y_1,...,y_N}$ and corresponding timestamps $d(y_i)$ for generating $y_i$,
the average lagging (AL) for a speech-to-text system is defined as
\begin{equation}
    \text{AL} = \frac{1}{\tau(T_\mathcal{X})} \sum^{\tau(T_\mathcal{X})}_{i=1} d(y_i) - \frac{T_\mathcal{X}}{|\mathbf{Y}|} \cdot (i - 1)
    \label{eq:al}
\end{equation}
Where the $\tau(T_x)$ is the first index of the target prediction that make use of the full source sentence.
The second term is the delay of an oracle system,
which starts the translation at the every beginning and stops exactly when the source sentence is finished.
To adapt AL to speech-to-speech task,
similar to \citet{zheng-etal-2020-fluent},
we first run an ASR system on the target speech $\mathcal{Y}(t)$ to generate $\mathbf{Y}$,
and then use a forced aligner to align $\mathbf{Y}$ with $\mathcal{Y}(t)$ to acquire $d(y_i), i={1,...,N}$.
To simulate a real world scenario in this paper,
we only consider computation-aware average lagging~\cite{ma2020simulmt},
in which $d(y_i)$ includes computation time.

\section{Background}
\subsection{Offline Direct S2ST with Direct Units}
Recently, \citet{lee2021direct} propose a method for offline direct S2ST with discrete units.
First, unsupervised continuous representations at every 20-ms frame are learned on target speech corpus by a HuBERT model \cite{hsu2021hubert}.
Then the k-means algorithm is applied to the representations to generate $K$ cluster centroids.
For each window, found the it's closest centroids index as it's discrete label. 
Denote the discrete sequence as $\mathbf{Z} = {z_1, ...z_L}$.
Given the discrete units,
\citet{lee2021direct} build a transformer-based~\cite{vaswani2017attention} speech-to-unit (S2U) model.
In the encoder, a stack of 1D-convolutional layers serve as downsampler for the speech input.
Since the target sequence is discrete, the S2U model can be trained with cross-entropy loss.
Finally, the vocoder converts the discrete units to speech signal.
\citet{lee2021direct} introduce a modified version of the HiFi-GAN neural vocoder~\cite{polyak2021high} for unit-to-waveform conversion.

\subsection{Monotonic Multihead Attention}
\label{sec:monotonic_attention}
Monotonic attention models
~\cite{raffel2017online,chiu2018mocha,arivazhagan-etal-2019-monotonic,ma2019monotonic}
are designed for the scenario where only partial input is provided to the decoder.
At a given time when $i-1$-th target token have been predicted
and $j$-th source inputs have been consumed,
a policy consisting in a stepwise probability $p_{i,j}$ predicts
whether to generate the $i$-th target token or to read the $j+1$-th source input.
Instead of sampling the actions from the policy,
\citet{raffel2017online} proposed an estimation of alignments between source and target
$\alpha_{i,j}$ from $p_{i,j}$ during training:
\begin{equation}
     \label{eq:monotonic_attention}
     \begin{split}
     \hat{\alpha}_{i,j} = p_{i,j} \left((1-p_{i,j-1})\frac{
         \hat{\alpha}_{i,j-1}}{p_{i,j-1}} + \alpha_{i-1, j}\right)\\
     \end{split}
\end{equation}
While \cite{raffel2017online} assume a hard alignment between encoder and decoder
(only one encoder state is passed to the decoder at any time),
\citet{chiu2018mocha} introduce monotonic chunkwise attention (MoChA),
which enables soft attention in a chunk following the moving attention head.
\citet{arivazhagan-etal-2019-monotonic} further extend (MoChA),
proposed monotonic infinite lookback attention (MILk),
in which a soft attention is computed over all the previous history.
Given the energy $u_{i,j}$ for the $i$-th decoder state and the $j$-th encoder state,
an expected soft attention is calculated in \autoref{eq:milk_recurent}:
\begin{equation}
    \label{eq:milk_recurent}
    \hat{\beta}_{i,j} = \sum_{k=j}^{|\mathbf{X}|} \left( \frac{\hat{\alpha}_{i, k} \exp(u_{i,j})}{\sum_{l=1}^k  \exp(u_{i, l})} \right)
\end{equation}
$\hat{\beta}$ instead of $\hat{\alpha}$ is then used in training.
\citet{arivazhagan-etal-2019-monotonic} also introduce latency augmented training for latency control,
Given the estimated alignment $a_{i,j}$,
the expected delays of all the predictions can be calculated.
A latency loss $\mathcal{C}([\hat{n}(y_1),...,\hat{n}(y_N)])$ is then added to the total loss function,
where $\mathcal{C}$ is a latency measurement function and $\hat{n}(y_i) = \sum_{j=1}^{|\mathbf{X}|} j \alpha_{i,j}$ is the expected delay of the system.
\citet{ma2019monotonic} further extend the monotonic attention to
monotonic multihead attention (MMA) for transformer models.
The design of MMA is to enable every attention head as individual monotonic attention.
Further, the latency is controlled by controlling the expected average and variance of the delays of all attention heads.

\subsection{Prefix-to-Prefix Policy (wait-$k$)}
A prefex-to-prefix (wait-$k$) policy~\cite{ma2019stacl} is rule-based policy
which read a certain amount of input and then read or write alternatively.
The initial look ahead, which is decide by $k$, determine the latency of the model.
\section{Methodology}
\subsection{Variational Monotonic Multihead Attention}
As introduced in \Cref{sec:monotonic_attention}, the alignment between source and target sequences is estimated as \Cref{eq:monotonic_attention}.
Such estimation shows good results when both source and target are short text sequences
However, \citet{ma2020simulmt} claimed that
such approach degrades the model performance on speech-to-text translation.
We also observed this deficient policy learnt
in the preliminary experiments on speech-to-speech task.
We found that as the length of speech sequence increases,
the estimation of alignment $\hat\alpha_{i,j}$ tends to be divergent,
which is due to the recurrent calculation of $\hat\alpha_{i,j}$ in \Cref{eq:monotonic_attention}. We provide a detailed analysis of such deficiency in appendix.
Therefore, a better estimation is needed for long sequence modeling.
However, such estimation is non-trivial,
since inference time alignment is not differentiable during training.

We propose variational monotonic multihead attention (V-MMA), which models the alignment with the latent variable $\alpha$ instead of recurrent estimation in sequence-to-sequence modeling.
Instead of optimizing the log probability $\log{p(Y|X})$, we optimize its evidence lower bound (ELBO) as below.
\begin{equation}
    \label{eq:elbo}
    \begin{aligned}
        \mathcal{L}(\omega, \phi, \theta) = &\mathbb{E}_{\alpha \sim q_{\phi}}\left[\log p_\theta(Y|X, \alpha)\right] \\
            & - \text{KL}\left[q_{\phi}(\alpha| Y, X) || p_\omega (\alpha|X)\right]\\
    \end{aligned}
\end{equation}
where $p_\omega (\alpha|X)$ is the priori of the monotonic alignment,
$q_{\phi}(\alpha| Y, X)$ is the posterior of the monotonic alignment
and $p_\theta(Y|X, \alpha)$ is the prediction of target sequence give input and alignment.
Since latent variable $\alpha$, in the form of a matrix, is a monotonic alignment between target and source sequences of size $|X|$ by $|Y|$,
we can find an one-to-one mapping between $\alpha$ and a binary sequence $Z$ of length $|X|+|Y|+1$,
where each $z \in Z$ indicates write action if $z=1$, otherwise read action.
Because the total number of write actions is the number of target words, we have $\sum_{z \in Z} z =|Y|$.
Sampling 1-D sequence of $Z$ is easier than 2-D matrix of $\alpha$.
In this paper, we want the policy to have less frequency of change for the purpose of smooth speech segment synthesis.
Therefore, we sample a sequence of the change of action $Z^*$ for a better control during the training time,
while for $1 < k < |X|+|Y|+1$
\begin{equation}
    z^*_k =
    \begin{cases}
        1 & z_{k-1} \ne z_{k}\\
        0 & \text{otherwise}
    \end{cases}
\end{equation}
During sampling, the each $z^*_k$ is sampled from a Bernoulli distribution parameterized by $p^*_k$,
\begin{equation}
    \label{eq:p}
    p^*_k = (1 - e^{\lambda (k-k')^2}) f(X_{:i}, Y_{:j}),
\end{equation}
Where $\lambda$ is a hyperparameter controling the frequency of change,
$i$ is the current target size,
$j$ is the current source size,
and $k'=\text{argmax}_{k'<k} \, Z^*_{k'}=1$ is the latest index of action change.
$f(X_{:i}, Y_{:j})$ is the context score, or equivalently $p_{i,j}$ in \Cref*{eq:monotonic_attention}.
Finally, the whole sequence of sampling $\alpha$ is $Z* \Rightarrow Z \Rightarrow \alpha$


Additionally, the calculation of the second term in \Cref{eq:elbo} can be written as
\begin{equation}
    \label{eq:KL}
    \begin{aligned}
        &\log \frac{q_{\phi}(\alpha| Y, X)}{p_\omega (\alpha|X)} \\
        &= \log \frac{\prod_{i=1}^{N} \phi_{i,z_i} \prod_{i=0}^{N} \prod_{j=a_{i}}^{z_{i+1}-1} (1 - \phi_{i,j})}{\prod_{i=1}^{N} \omega_{i,z_i} \prod_{i=0}^{N} \prod_{j=z_{i}}^{z_{i+1}-1} (1 - \omega_{i,j})}\\
        &= \sum_{i=1}^N \log \frac{\phi_{i,z_i}}{\omega_{i,z_i}} + \sum_{i=1}^N \sum_{j=z_i}^{z_{i+1}-1} \log \frac{1 - \phi_{i,j}}{1 - \omega_{i,j}}
    \end{aligned}
\end{equation}
Where the $\phi$ is the posterior probability matrix and $\omega$ is the priori probability matrix.
We explore two types of priori, one is a fixed diagonal based priori,
the other one is an offline label priori introduced in \Cref{section:policy}


%

\subsection{Simultaneous Speech-to-Units Model}
\label{section:policy}
\begin{figure}[h]
    \includegraphics[width=0.5\textwidth]{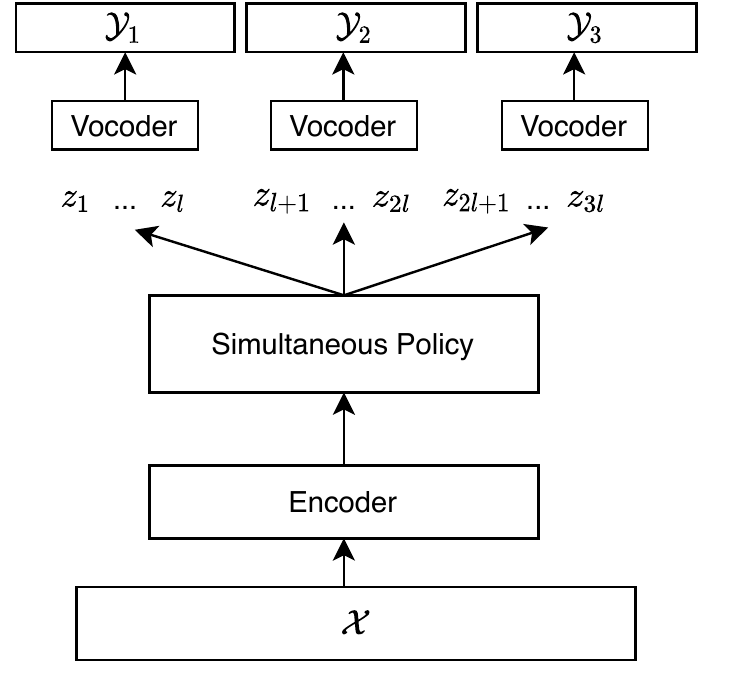}
    \caption{Architecture of direct Simul-S2ST model with discrete units.}
    \label{fig:arch}
\end{figure}

\Cref{fig:arch} illustrate the architecture of the direct Simul-S2ST model with discrete units.
The encoder reads the speech features with downsampling,
and generates a sequence of hidden representations.
Then the simultaneous policy replaces soft-attention in offline model to connect encoder and decoder.
Because of the granularity of the encoder states,
we follow the same setup as \cite{ma2020simulmt} to use a fixed pre-decision module before applying the simultaneous policy.
The policy contains two actions: \textsc{READ} and \textsc{WRITE}.
The \textsc{READ} action indicates that the model takes another chunk of speech segment to update the encoder states, while the \textsc{WRITE} action predicts the discrete units. 

A vocoder will be applied to the discrete units to synthesize the final speech output.
Denote the emission rate $l$, the vocoder will be called every time $l$ units are predicted.
Note that the vocoder is not used during training time.
Because the vocoder is trained on short speech segment in a non-autoregressive manner,
as shown in the results, a small $l$ can achieve good latency without a huge sacrifice on quality.

\subsection{Simultaneous Policy with Supervision}

A challenge of simultaneous translation is to learn a good policy that decides when to read more input and when to generate output.
In this work, we propose to distill the policy knowledge from an offline model into the simultaneous model.
The intuition is that a well-trained offline model captures the alignment between inputs and outputs.

\noindent \textbf{Offline Policy}.
Partial inputs with different lengths are provided, we could know how much input is needed for each target prediction.
Given a trained offline model, we want to export its learned policy, i.e.,
how much input information is needed to generate a target output.
We provide a partial input with which the offline model predicts the likelihood of each target token in the reference translation.
The target token is ranked among all tokens in the vocabulary based on the estimated likelihood.
Given a rank threshold $r$, if a token has a higher ranker than $r$, it suggests that the partial input contains sufficient information for this token. Hence a simultaneous model could generate the token when reading that partial input.

With the policy obtained from the offline model, we want to guide the simultaneous model using knowledge distillation.

\noindent \textbf{Knowledge distillation in cascaded Simul-S2S}. For cascaded Simul-S2S model, we integrate the offline knowledge into MMA in the simultaneous S2T component. This is done via the supervision on the simultaneous model's attention.
Suppose that the lengths of partial inputs sorted in an increasing order $[L_{1}, \ldots, L_{i}, \ldots]$. For the $j$-th token, an offline model ranks it above the threshold $r$ when the partial input of length $L_{i}$ is given. Suppose that the attention vector of the target token $j$ over the sequence of encoder states is $\boldsymbol{\alpha}_{j}=[\alpha_{j,1}, \ldots, \alpha_{j,i}, \ldots]$.

A partial input of length $L_{i-1}$ does not provide enough information to predict token $j$ while an input of length $L_{i}$ does. This implies that the source knowledge that token $j$ relies on fall in the input segment $(L_{i-1}+1, L_{i}]$. Therefore we encourage the attention of a simultaneous model to be concentrated on this segment. Accordingly, the auxiliary loss $L_{a}$ we add to the simultaneous S2T training is:
\begin{align}
\label{eq:aux_loss}
L_{a} = -\sum\limits_{k=L_{i-1}+1}^{L_{i}}\alpha_{j,k}.
\end{align}

The auxiliary loss is added to the training objective as complimentary to the prediction loss and latency loss of MMA.

\noindent \textbf{Knowledge distillation in direct Simul-S2S}. For direct Simul-S2S model, we leverage the offline knowledge from the direct offline S2S model into V-MMA. We find the alignment matrix between source and target sequences using an offline model as above.
The alignment from offline policy is then normalized along the target direction, and used as $\omega$ in \Cref{eq:KL}.
\section{Related Work}

\textbf{Speech-to-speech translation}.
The task of speech-to-speech translation translates speech in one language into speech in another.
Earlier approaches are cascaded models consisting of automatic speech recognition (ASR),
machine translation (MT) and text-to-speech synthesis (TTS).
Recent studies make progress in direct S2S translation without relying on intermediate texts.
In comparison with cascaded models, direct translation is faster in inference and free from error propagation in the pipeline.
Translatron is a sequence-to-sequence model with a multitask objective to predict source and target transcripts as well as generating spectrograms of the target speech \cite{jia2019direct}.
Translatron2 is further proposed with improvements in voice preservation, speech naturalness and prediction robustness over Translatron \cite{jia2021translatotron}.
Instead of modeling spectrograms of the target speech, \citet{lee2021direct} predicts discrete units which are learned from speech corpora in an self-supervised manner in its multitask learning framework for S2S training \cite{lee2021direct}.

\noindent\textbf{Simultaneous translation}. Simultaneous translation equips the model with the capability of live translation, generating outputs before it finishes reading complete inputs. A challenge in simultaneous translation is to find a good policy deciding when to start translation based on partial inputs, taking both the translation quality and latency into consideration.

Rule-based policies are used for reading and writing such as wait-if-diff and wait-if-worse policy \cite{cho2016can}, and tunable agent policy \cite{dalvi2018incremental}. A simple yet efficient wait-$k$ strategy is proposed so that the model reads first $k$ inputs, and then alternate reads and writes with a given step \cite{ma2019stacl}. In a simultaneous S2T model, speech segmenter splits the input streaming speech and the encoder-decoder attention adopts a wait-$k$ strategy \cite{ren2020simulspeech}. 

Learned policies have attracted increasing attention due to their flexibility to balance the tradeoff between translation quality and latency. Monotonic multihead attention (MMA) is a variant of multihead attention, enabling the model in partial input processing and incremental decoding \cite{ma2019monotonic}.
An adaptive policy is designed via a heuristic composition of a set of fixed policies for a simultaneous S2T model \cite{zheng2020simultaneous}.

\begin{table*}[htbp!]
\centering
\begin{tabular}{|c|c|c|c|c|}
\hline
 & \multicolumn{2}{c|}{} & BLEU & CA AL ($ms$) \\ \hline
\multirow{2}{*}{Offline} & \multicolumn{2}{c|}{Cascaded (S2T+TTS)} & 39.5 & -  \\ \cline{2-5}
 & \multicolumn{2}{c|}{Direct S2U} & 37.2 & -  \\ \hline
\multirow{16}{*}{Simul} & \multirow{7}{*}{\begin{tabular}[c]{@{}c@{}}Cascaded \\ (S2T with wait-$k$ \\ + incremental TTS)\end{tabular}} & $k$=1 & 23.9 &   1317 \\ \cline{3-5}
 &  & $k$=3 & 25.4 & 3109 \\ \cline{3-5}
 &  & $k$=5 & 25.9 &  3004 \\ \cline{3-5}
 &  & $k$=10 & 28.1 &  3476  \\ \cline{3-5}
 &  & $k$=15 & 29.2 & 3803 \\ \cline{3-5}
 &  & $k$=20 & 30.1  &  4099  \\ \cline{3-5}
 &  & $k$=25 & 31.4 & 4460  \\ \cline{2-5}
 & \multirow{3}{*}{\begin{tabular}[c]{@{}c@{}}Cascaded \\ (S2T with MMA \\ + incremental TTS)\end{tabular}}
 & lw=$1e-4$ & 28.4 & 3884 \\\cline{3-5}
 &  & lw=$1e-3$ & 30.5 & 3737  \\ \cline{3-5}
 &  & lw=$5e-4$ & 35.8 & 4504  \\ \cline{2-5}
 & \multirow{2}{*}{\begin{tabular}[c]{@{}c@{}}Cascaded (S2T with MMA \\ + offline policy + incremental TTS)\end{tabular}}
 &  lw=$1e-3$, aw=$0..1$ & 28.8 & 3834  \\ \cline{3-5}
 & & lw=$1e-4$, aw=$0.1$ & 36 & 3551 \\\cline{2-5}
 & \multirow{3}{*}{Direct S2U with wait-$k$}
 & $k$=5 & 22.2 & 1757 \\\cline{3-5}
 &  & $k$=10 & 27.9  &  3520  \\ \cline{3-5}
 &  & $k$=15 & 33.5 &  4127 \\ \cline{3-5}
 &  & $k$=20 & 34.3 &   4409  \\ \cline{2-5}
 & \multirow{3}{*}{Direct S2U with V-MMA}
 &  $\lambda$=0.01, w/o label& 25.3 & 3136 \\\cline{3-5}
 & & $\lambda$=0.01, w/ label& 25.9  & 3215 \\ \cline{3-5}
 & & $\lambda$=0.5, w/o label & 33.4 &  4564 \\ \cline{3-5}
 & & $\lambda$=0.5, w/ label & 34 &  4558  \\ \hline
\end{tabular}
\caption{BLEU scores and latency of models on the Fisher Spanish-English dataset.}

\label{tab:fisher_results}
\end{table*}
\begin{table*}[htbp!]
\centering
\begin{tabular}{|c|c|c|c|c|}
\hline
 & \multicolumn{2}{c|}{} & BLEU & CA AL ($ms$) \\ \hline
\multirow{2}{*}{Offline} & \multicolumn{2}{c|}{Cascaded (S2T+TTS)} & 24.4 & -  \\ \cline{2-5}
 & \multicolumn{2}{c|}{Direct S2U} & 23.7 & -  \\ \hline
\multirow{19}{*}{Simul} & \multirow{4}{*}{\begin{tabular}[c]{@{}c@{}}Cascaded \\ (S2T with wait-$k$ \\ + incremental TTS)\end{tabular}}
 & $k$=3 & 9.7 & 1094  \\ \cline{3-5}
 &  & $k$=5 & 11.7 & 2892 \\ \cline{3-5}
 &  & $k$=7 & 12.8 & 3108  \\ \cline{3-5}
 &  & $k$=9 & 14.3 & 3159  \\ \cline{2-5}
& \multirow{2}{*}{\begin{tabular}[c]{@{}c@{}}Cascaded (S2T with MMA \\ + incremental TTS)\end{tabular}} & lw=$1.5e-3$ & 15.8  & 3249  \\ \cline{3-5}
 &  & lw=$1e-3$ & 15.9 & 3283 \\ \cline{3-5}
 &  & lw=$1.2e-3$ & 15.9  & 3303  \\  \cline{3-5}
 &  & lw=$8e-4$ & 16.4  & 3547  \\  \cline{2-5}
& \multirow{2}{*}{\begin{tabular}[c]{@{}c@{}}Cascaded (S2T with MMA \\ + offline policy + incremental TTS)\end{tabular}}  & lw=$1.5e-3$, aw=$0.1$ & 15.8 & 3991 \\ \cline{3-5}
 &  & lw=$1e-3$, , aw=$0.1$ & 16.0 & 3716 \\ \cline{3-5}
 & & lw=$8e-4$, , aw=$0.1$ & 16.6 & 3743 \\ \cline{2-5}
 & \multirow{3}{*}{Direct S2U with wait-$k$}
 & $k$=5 & 9.2 & 839 \\\cline{3-5}
 &  & $k$=10 & 10.8 & 2718 \\ \cline{3-5}
 &  & $k$=15 & 16.7 & 3278  \\ \cline{3-5}
 &  & $k$=20 & 18.6 & 4473   \\ \cline{2-5}
 & \multirow{3}{*}{Direct S2U with V-MMA}
 &  $\lambda$=0.01, w/o label& 11.5 & 2706 \\\cline{3-5}
 & & $\lambda$=0.01, w/ label& 12.2  & 2927 \\ \cline{3-5}
 & & $\lambda$=0.5, w/o label & 18.1 &  4421 \\ \cline{3-5}
 & & $\lambda$=0.5, w/ label & 18.2 &  4534 \\ \hline
\end{tabular}
\caption{BLEU scores and latency of models on the MuST-C English-Spanish dataset.}
\label{tab:mustc_results}
\end{table*}

\section{Experiments}
In this section, we conduct experiments on simultaneous speech-to-speech translation.

\noindent \textbf{Dataset}.
Two datasets--Fisher and MuST-C--commonly used for speech translation tasks are used in this work.
\begin{itemize}
    \item Fisher Spanish-English dataset \cite{post2014fisher}. The dataset consists of $139k$ sentences from telephone conversations in Spanish, the corresponding Spanish text transcriptions and their English text translation. As \cite{lee2021direct}, a high-quality in-house TTS is used to prepare English speech with a single female voice.
    \item MuST-C dataset \cite{cattoni2021must}. It is a multilingual speech translation corpus collected from TED Talks. We use English-Spanish data, where we synthesize Spanish speech from Spanish texts provided by MuST-C with the help of an in-house TTS model.
\end{itemize}

\noindent \textbf{Evaluation metrics}. Translation quality and efficiency are both important for simultaneous translation models. Following the evaluation of speech quality~\cite{lee2021direct}, we transcribe the generated speech with a pre-trained ASR model, and measure the BLEU score by comparing the transcription and the reference.
We keep discontinuities in simultaneously generated speech.

The computation-aware~\cite{ma2020simulmt} version of Average Lagging (AL)~\cite{ma-etal-2019-stacl}
is used to measure the latency as previous works of simultaneous translation.

\noindent \textbf{Baselines}. Various models are included for a comprehensive empirical comparison.
\begin{itemize}
    \item Offline cascaded S2S translation. The offline cascaded system consists of two components: S2T and TTS. The S2T model is the \text{s2t\_transformer\_s} architecture provided by \textsc{FAIRSEQ} S2T for speech-to-text translation \cite{wang2020fairseq}. The target vocabulary in S2T model consists of $47$ English characters on Fisher data, and $8000$ Spanish unigrams on MuST-C data.
    As for the TTS, we use a Transformer model with $6$ layers, $4$ attention heads and dimensions of $512$ and $2048$, and a HiFi-GAN based vocoder generating speech from the predicted mel-spectrograms \cite{kong2020hifi}.
    \item Simultaneous cascaded S2S model consisting of S2T and TTS modules. Its S2T component has the same architecture as the offline S2T Transformer. As for the simultaneous strategies in S2T part, we use three methods: (1) Wait-$k$ strategy, (2) MMA strategy, and (3) MMA strategy with the supervision of offline policy. Each strategy could control the translation latency with its own hyperparameters. Wait-$k$ strategy tunes the value of $k$, MMA strategy sets the latency weight $\text{lw}$ in training objective, and larger $lw$ reduces latency. As for MMA combined with offline policy, we assign weight $\text{aw}$ to the auxiliary loss in Eq.(\ref{eq:aux_loss}) besides $\text{lw}$ on latency loss.
    As for the incremental TTS module in the cascaded model, we adapt the non-autoregressive FastSpeech 2 model \cite{fastspeech2}. It incrementally generates utterances word-by-word, utilizing the duration predictor for segmenting output word boundaries \cite{stephenson21_interspeech}. It further uses a lookahead of 1 word, which is equivalent to a wait-$k$ strategy at test-time \cite{ma2019stacl,ma-etal-2020-incremental}, where $k$ is two words\footnote{Wait-$(k+1)$ is equivalent to lookahead-$k$ following the definition in \cite{ma-etal-2020-incremental}}.
    To improve performance on partial inputs, we apply the prefix augmentation procedure described in \cite{liu2021incremental}. 
    \item Offline direct S2S translation \cite{lee2021direct}. It is a state-of-the-art direct translation model without reliance on intermediate text outputs. On Fisher dataset, the direct model uses $12$ encoder layers, $6$ decoder layers, $4$ attention heads, an embedding dimension of $256$ and a feedforward dimension of $2048$. As for MuST-C data, we use a larger model with attention heads increased to $8$ and the embedding dimension increased to $512$.
\end{itemize}

Offline models would provide an upper bound of the translation quality. As the first work in simultaneous speech-to-speech translation, we explore various simultaneous strategies including wait-$k$ strategy, V-MMA strategy with and without offline knowledge in the direct S2U translation model. We follow the same training setup as \citet{lee2021direct},
except that a masked attention for our simultaneous policies is used for mode training with partial inputs \cite{ma-etal-2019-stacl}.
As to inference, we use SemEval toolkit to simulate the real-world simultaneous translation setup with a server-client scheme \cite{ma2020simuleval}.

\section{Results}
\subsection{Fisher}
The results on Fisher dataset are shown in \Cref{tab:fisher_results}.

\textbf{Cascaded models}. Wait-$k$ and MMA strategies demonstrate comparable performance in terms of BLEU and latency. When we compare MMA strategy with and without offline knowledge, offline knowledge brings empirical gains, e.g., it has a BLEU score of $36$ (v. $30.5$ w/o offline knowledge) at the latency of $3551$ ms (v. $3747$ ms w/o offline knowledge).

\textbf{Direct models}. When it comes to direct models, the V-MMA policy controls mdoel by the hyperparameter $\lambda$ in \Cref{eq:p}. Offline knowledge brings gains of $0.6$ BLEU to V-MMA with a comparable latency. V-MMA policy with offline labels has similar performance compared with wait-$k$ policy in direct S2S models.

\textbf{Direct v. cascaded}. We include offline speech-to-speech models which provide an upper bound of the translation quality. In the offline setting, the direct S2S model falls behind the cascaded model by $2.3$ BLEU.
As for the simultaneous setting, latency is an important factor besides the translation quality.
With wait-$k$ policy, direct models achieve higher BLEU than cascaded models in the high-latency setting ($>4000$ ms)
This demonstrates that direct models have the ability to handle compounding errors while the error propagation hurts the performance of cascaded models.
In the low-latency region ($<4000$ ms), direct models have comparable BLEU and latency in comparsion with cascaded models when wait-$k$ strategy is applied.

\subsection{MuST-C}
We report model results on MuST-C English-Spanish dataset in \Cref{tab:mustc_results}.

\textbf{Simultaneous policies}. For the direct simultaneous model, 
V-MMA outperforms wait-$k$ given latency lower than $3000$ ms, e.g., V-MMA has a BLEU of $11.5$ which is higher than wait-$k$'s BLEU of $10.8$ with a latency of $2706$ ms which is comparable to wait-$k$'s latency of $2718$ ms. When it comes to high-latency region, V-MMA has a lower BLEU of $18.1$ with $4421$-ms latency than wait-$k$ which has a BLEU of $18.6$ with $4473$-ms latency.
The integration of offline knowledge increases the latency of MMA in cascaded models with similar BLEU scores, and it shows slight improvements to V-MMA in direct models. 

\textbf{Direct v. cascaded}. We again include offline speech-to-speech models which provide an upper bound of the translation quality. In the offline setting, the direct model falls behind the cascaded model by $0.7$ BLEU. In the simultaneous setting, both wait-$k$ and V-MMA in direct models have higher BLEU scores in comparison with wait-$k$ and MMA in cascaded models given latency higher than $3000$ ms.


\subsection{Discontinuity in Speech Synthesis}
\label{sec:discont}
In cascaded Simul-S2S model, speech segments are generated incrementally by TTS, and we add discontinuity between segments to mimic the real-world scenario.
\Cref*{tab:fisher-discont} and \Cref*{tab:mustc-discont} report the loss of BLEU scores due to the discontinuity for cascaded models with waitk-$k$ policy.
\begin{table}[htbp!]
\centering
\begin{tabular}{|c|c|c|c|}
\hline
$k$ & w/ discont & w/o discont & loss \\
\hline
1& 30.1& 23.9& 6.2 \\
3& 31.8& 25.4& 6.4\\
5& 32.4& 25.9& 6.5\\
10& 34.5& 28.1& 6.4\\
15& 34.8& 29.2& 5.6\\
25& 35.7& 31.4& 4.3\\
\hline
& & & Average = 5.4\\
\hline
\end{tabular}
\caption{BLEU score loss for wait-$k$ policy on Fisher dataset.}
\label{tab:fisher-discont}
\end{table}
\begin{table}[htbp!]
\centering
\begin{tabular}{|c|c|c|c|}
\hline
$k$ & w/ discont & w/o discont & loss \\
\hline
3 & 14.4& 9.7& 4.7 \\
5 & 17.1& 11.7& 5.4 \\
7 & 18.7& 12.8& 5.9 \\
9 & 19.4& 14.3& 5.1 \\
11 & 20.3& 14.4& 5.9 \\
13 & 21.2& 15.1& 6.1 \\
15 & 21.4& 16.3& 5.1 \\
17 & 22& 17.3& 4.7 \\
\hline
& & & Average=5.9\\
\hline
\end{tabular}
\caption{BLEU score loss for wait-$k$ policy on MuST-C dataset.}
\label{tab:mustc-discont}
\end{table}
We can observe that there are a significant and consistent quality drop after considering the discontinuity when synthesizing the speech.
Another factor of such drop is that the ASR system we used for evaluation is trained on continuous speech,
so that a worse transcripts of target speech is used for evaluation.
This suggests that for further evaluation, we should consider human evaluation not only on mean option score on naturalness, but also on the quality of the translation.
\section{Conclusion}
In this work, we propose a direct simultaneous speech-to-speech translation model.
We then propose a V-MMA model for speech-to-speech scenario.
We further carry out experiments on Fisher and MuST-C dataset,
and show that the direct models is compared with cascaded approach on high latency setting but worse on low latency setting.
In the future, we plan to carry out human evaluation on both naturalness and quality of the speech output.

\bibliography{bibs/asr,bibs/dataset,bibs/mt,bibs/simul,bibs/st,bibs/toolkit,bibs/tts}

\begin{thebibliography}{30}
\providecommand{\natexlab}[1]{#1}
\providecommand{\url}[1]{\texttt{#1}}
\expandafter\ifx\csname urlstyle\endcsname\relax
  \providecommand{\doi}[1]{doi: #1}\else
  \providecommand{\doi}{doi: \begingroup \urlstyle{rm}\Url}\fi

\bibitem[Arivazhagan et~al.(2019)Arivazhagan, Cherry, Macherey, Chiu, Yavuz,
  Pang, Li, and Raffel]{arivazhagan-etal-2019-monotonic}
Arivazhagan, N., Cherry, C., Macherey, W., Chiu, C.-C., Yavuz, S., Pang, R.,
  Li, W., and Raffel, C.
\newblock Monotonic infinite lookback attention for simultaneous machine
  translation.
\newblock In \emph{Proceedings of the 57th Annual Meeting of the Association
  for Computational Linguistics}, pp.\  1313--1323, Florence, Italy, July 2019.
  Association for Computational Linguistics.

\bibitem[Cattoni et~al.(2021)Cattoni, Di~Gangi, Bentivogli, Negri, and
  Turchi]{cattoni2021must}
Cattoni, R., Di~Gangi, M.~A., Bentivogli, L., Negri, M., and Turchi, M.
\newblock Must-c: A multilingual corpus for end-to-end speech translation.
\newblock \emph{Computer Speech \& Language}, 66:\penalty0 101155, 2021.

\bibitem[Chiu \& Raffel(2018)Chiu and Raffel]{chiu2018mocha}
Chiu, C.-C. and Raffel, C.
\newblock Monotonic chunkwise attention.
\newblock 2018.
\newblock URL \url{https://openreview.net/pdf?id=Hko85plCW}.

\bibitem[Cho \& Esipova(2016)Cho and Esipova]{cho2016can}
Cho, K. and Esipova, M.
\newblock Can neural machine translation do simultaneous translation?
\newblock \emph{arXiv preprint arXiv:1606.02012}, 2016.

\bibitem[Dalvi et~al.(2018)Dalvi, Durrani, Sajjad, and
  Vogel]{dalvi2018incremental}
Dalvi, F., Durrani, N., Sajjad, H., and Vogel, S.
\newblock Incremental decoding and training methods for simultaneous
  translation in neural machine translation.
\newblock In \emph{NAACL-HLT (2)}, 2018.

\bibitem[Gu et~al.(2017)Gu, Neubig, Cho, and Li]{gu2017learning}
Gu, J., Neubig, G., Cho, K., and Li, V.~O.
\newblock Learning to translate in real-time with neural machine translation.
\newblock In \emph{15th Conference of the European Chapter of the Association
  for Computational Linguistics, EACL 2017}, pp.\  1053--1062. Association for
  Computational Linguistics (ACL), 2017.

\bibitem[Hsu et~al.(2021)Hsu, Bolte, Tsai, Lakhotia, Salakhutdinov, and
  Mohamed]{hsu2021hubert}
Hsu, W.-N., Bolte, B., Tsai, Y.-H.~H., Lakhotia, K., Salakhutdinov, R., and
  Mohamed, A.
\newblock Hubert: Self-supervised speech representation learning by masked
  prediction of hidden units.
\newblock \emph{arXiv preprint arXiv:2106.07447}, 2021.

\bibitem[Jia et~al.(2019)Jia, Weiss, Biadsy, Macherey, Johnson, Chen, and
  Wu]{jia2019direct}
Jia, Y., Weiss, R.~J., Biadsy, F., Macherey, W., Johnson, M., Chen, Z., and Wu,
  Y.
\newblock Direct speech-to-speech translation with a sequence-to-sequence
  model.
\newblock \emph{arXiv preprint arXiv:1904.06037}, 2019.

\bibitem[Jia et~al.(2021)Jia, Ramanovich, Remez, and
  Pomerantz]{jia2021translatotron}
Jia, Y., Ramanovich, M.~T., Remez, T., and Pomerantz, R.
\newblock Translatotron 2: Robust direct speech-to-speech translation.
\newblock \emph{arXiv preprint arXiv:2107.08661}, 2021.

\bibitem[Kong et~al.(2020)Kong, Kim, and Bae]{kong2020hifi}
Kong, J., Kim, J., and Bae, J.
\newblock {H}i{F}i-{GAN}: Generative adversarial networks for efficient and
  high fidelity speech synthesis.
\newblock \emph{Advances in Neural Information Processing Systems}, 33, 2020.

\bibitem[Lee et~al.(2021)Lee, Chen, Wang, Gu, Ma, Polyak, Adi, He, Tang, Pino,
  et~al.]{lee2021direct}
Lee, A., Chen, P.-J., Wang, C., Gu, J., Ma, X., Polyak, A., Adi, Y., He, Q.,
  Tang, Y., Pino, J., et~al.
\newblock Direct speech-to-speech translation with discrete units.
\newblock \emph{arXiv preprint arXiv:2107.05604}, 2021.

\bibitem[Liu et~al.(2021)Liu, Wang, Gong, Ma, Tang, and
  Pino]{liu2021incremental}
Liu, D., Wang, C., Gong, H., Ma, X., Tang, Y., and Pino, J.
\newblock Incremental speech synthesis for speech-to-speech translation.
\newblock \emph{arXiv preprint arXiv:2110.08214}, 2021.

\bibitem[Ma et~al.(2019{\natexlab{a}})Ma, Huang, Xiong, Zheng, Liu, Zheng,
  Zhang, He, Liu, Li, Wu, and Wang]{ma-etal-2019-stacl}
Ma, M., Huang, L., Xiong, H., Zheng, R., Liu, K., Zheng, B., Zhang, C., He, Z.,
  Liu, H., Li, X., Wu, H., and Wang, H.
\newblock {STACL}: Simultaneous translation with implicit anticipation and
  controllable latency using prefix-to-prefix framework.
\newblock In \emph{Proceedings of the 57th Annual Meeting of the Association
  for Computational Linguistics}, pp.\  3025--3036, Florence, Italy, July
  2019{\natexlab{a}}. Association for Computational Linguistics.

\bibitem[Ma et~al.(2019{\natexlab{b}})Ma, Huang, Xiong, Zheng, Liu, Zheng,
  Zhang, He, Liu, Li, et~al.]{ma2019stacl}
Ma, M., Huang, L., Xiong, H., Zheng, R., Liu, K., Zheng, B., Zhang, C., He, Z.,
  Liu, H., Li, X., et~al.
\newblock Stacl: Simultaneous translation with implicit anticipation and
  controllable latency using prefix-to-prefix framework.
\newblock In \emph{Proceedings of the 57th Annual Meeting of the Association
  for Computational Linguistics}, pp.\  3025--3036, 2019{\natexlab{b}}.

\bibitem[Ma et~al.(2020{\natexlab{a}})Ma, Zheng, Liu, Zheng, Liu, Peng, Church,
  and Huang]{ma-etal-2020-incremental}
Ma, M., Zheng, B., Liu, K., Zheng, R., Liu, H., Peng, K., Church, K., and
  Huang, L.
\newblock Incremental text-to-speech synthesis with prefix-to-prefix framework.
\newblock In \emph{Findings of the Association for Computational Linguistics:
  EMNLP 2020}, pp.\  3886--3896, Online, November 2020{\natexlab{a}}.
  Association for Computational Linguistics.
\newblock \doi{10.18653/v1/2020.findings-emnlp.346}.
\newblock URL \url{https://aclanthology.org/2020.findings-emnlp.346}.

\bibitem[Ma et~al.(2019{\natexlab{c}})Ma, Pino, Cross, Puzon, and
  Gu]{ma2019monotonic}
Ma, X., Pino, J.~M., Cross, J., Puzon, L., and Gu, J.
\newblock Monotonic multihead attention.
\newblock In \emph{International Conference on Learning Representations},
  2019{\natexlab{c}}.

\bibitem[Ma et~al.(2020{\natexlab{b}})Ma, Dousti, Wang, Gu, and
  Pino]{ma2020simuleval}
Ma, X., Dousti, M.~J., Wang, C., Gu, J., and Pino, J.
\newblock Simuleval: An evaluation toolkit for simultaneous translation.
\newblock \emph{arXiv preprint arXiv:2007.16193}, 2020{\natexlab{b}}.

\bibitem[Ma et~al.(2020{\natexlab{c}})Ma, Pino, Cross, Puzon, and
  Gu]{ma2020simulmt}
Ma, X., Pino, J., Cross, J., Puzon, L., and Gu, J.
\newblock {SimulMT to SimulST: Adapting Simultaneous Text Translation to
  End-to-End Simultaneous Speech Translation}.
\newblock In \emph{Proceedings of 2020 Asia-Pacific Chapter of the Association
  for Computational Linguistics and the International Joint Conference on
  Natural Language Processing}, 2020{\natexlab{c}}.

\bibitem[Ma et~al.(2021)Ma, Wang, Dousti, Koehn, and Pino]{ma2021streaming}
Ma, X., Wang, Y., Dousti, M.~J., Koehn, P., and Pino, J.
\newblock Streaming simultaneous speech translation with augmented memory
  transformer.
\newblock In \emph{ICASSP 2021-2021 IEEE International Conference on Acoustics,
  Speech and Signal Processing (ICASSP)}, pp.\  7523--7527. IEEE, 2021.

\bibitem[Polyak et~al.(2021)Polyak, Wolf, Adi, Kabeli, and
  Taigman]{polyak2021high}
Polyak, A., Wolf, L., Adi, Y., Kabeli, O., and Taigman, Y.
\newblock High fidelity speech regeneration with application to speech
  enhancement.
\newblock In \emph{ICASSP 2021-2021 IEEE International Conference on Acoustics,
  Speech and Signal Processing (ICASSP)}, pp.\  7143--7147. IEEE, 2021.

\bibitem[Post et~al.(2014)Post, Kumar, Lopez, Karakos, Callison-Burch, and
  Khudanpur]{post2014fisher}
Post, M., Kumar, G., Lopez, A., Karakos, D., Callison-Burch, C., and Khudanpur,
  S.
\newblock Fisher and callhome spanish--english speech translation, 2014.

\bibitem[Raffel et~al.(2017)Raffel, Luong, Liu, Weiss, and
  Eck]{raffel2017online}
Raffel, C., Luong, M.-T., Liu, P.~J., Weiss, R.~J., and Eck, D.
\newblock Online and linear-time attention by enforcing monotonic alignments.
\newblock In \emph{Proceedings of the 34th International Conference on Machine
  Learning-Volume 70}, pp.\  2837--2846. JMLR. org, 2017.

\bibitem[Ren et~al.(2020)Ren, Liu, Tan, Zhang, Tao, Zhao, and
  Liu]{ren2020simulspeech}
Ren, Y., Liu, J., Tan, X., Zhang, C., Tao, Q., Zhao, Z., and Liu, T.-Y.
\newblock Simulspeech: End-to-end simultaneous speech to text translation.
\newblock In \emph{Proceedings of the 58th Annual Meeting of the Association
  for Computational Linguistics}, pp.\  3787--3796, 2020.

\bibitem[Ren et~al.(2021)Ren, Hu, Tan, Qin, Zhao, Zhao, and Liu]{fastspeech2}
Ren, Y., Hu, C., Tan, X., Qin, T., Zhao, S., Zhao, Z., and Liu, T.
\newblock Fastspeech 2: Fast and high-quality end-to-end text to speech.
\newblock In \emph{9th International Conference on Learning Representations,
  {ICLR} 2021, Virtual Event, Austria, May 3-7, 2021}. OpenReview.net, 2021.
\newblock URL \url{https://openreview.net/forum?id=piLPYqxtWuA}.

\bibitem[Stephenson et~al.(2021)Stephenson, Hueber, Girin, and
  Besacier]{stephenson21_interspeech}
Stephenson, B., Hueber, T., Girin, L., and Besacier, L.
\newblock {Alternate Endings: Improving Prosody for Incremental Neural TTS with
  Predicted Future Text Input}.
\newblock In \emph{Proc. Interspeech 2021}, pp.\  3865--3869, 2021.
\newblock \doi{10.21437/Interspeech.2021-275}.

\bibitem[Sudoh et~al.(2020)Sudoh, Kano, Novitasari, Yanagita, Sakti, and
  Nakamura]{sudoh2020simultaneous}
Sudoh, K., Kano, T., Novitasari, S., Yanagita, T., Sakti, S., and Nakamura, S.
\newblock Simultaneous speech-to-speech translation system with neural
  incremental asr, mt, and tts.
\newblock \emph{arXiv preprint arXiv:2011.04845}, 2020.

\bibitem[Vaswani et~al.(2017)Vaswani, Shazeer, Parmar, Uszkoreit, Jones, Gomez,
  Kaiser, and Polosukhin]{vaswani2017attention}
Vaswani, A., Shazeer, N., Parmar, N., Uszkoreit, J., Jones, L., Gomez, A.~N.,
  Kaiser, {\L}., and Polosukhin, I.
\newblock Attention is all you need.
\newblock In \emph{Advances in neural information processing systems}, pp.\
  5998--6008, 2017.

\bibitem[Wang et~al.(2020)Wang, Tang, Ma, Wu, Okhonko, and
  Pino]{wang2020fairseq}
Wang, C., Tang, Y., Ma, X., Wu, A., Okhonko, D., and Pino, J.
\newblock Fairseq s2t: Fast speech-to-text modeling with fairseq.
\newblock In \emph{Proceedings of the 1st Conference of the Asia-Pacific
  Chapter of the Association for Computational Linguistics and the 10th
  International Joint Conference on Natural Language Processing: System
  Demonstrations}, pp.\  33--39, 2020.

\bibitem[Zheng et~al.(2020{\natexlab{a}})Zheng, Liu, Zheng, Ma, Liu, and
  Huang]{zheng2020simultaneous}
Zheng, B., Liu, K., Zheng, R., Ma, M., Liu, H., and Huang, L.
\newblock Simultaneous translation policies: From fixed to adaptive.
\newblock In \emph{Proceedings of the 58th Annual Meeting of the Association
  for Computational Linguistics}, pp.\  2847--2853, 2020{\natexlab{a}}.

\bibitem[Zheng et~al.(2020{\natexlab{b}})Zheng, Ma, Zheng, Liu, Yuan, Church,
  and Huang]{zheng-etal-2020-fluent}
Zheng, R., Ma, M., Zheng, B., Liu, K., Yuan, J., Church, K., and Huang, L.
\newblock Fluent and low-latency simultaneous speech-to-speech translation with
  self-adaptive training.
\newblock In \emph{Findings of the Association for Computational Linguistics:
  EMNLP 2020}, pp.\  3928--3937, Online, November 2020{\natexlab{b}}.
  Association for Computational Linguistics.
\newblock \doi{10.18653/v1/2020.findings-emnlp.349}.
\newblock URL \url{https://aclanthology.org/2020.findings-emnlp.349}.

\end{thebibliography}
\bibliographystyle{icml2022}
\end{document}